\documentclass[conference]{IEEEtran}

\usepackage[dvips]{graphicx}
\usepackage{amsmath,amssymb}
\usepackage{algorithm}
\usepackage{algorithmic}
\usepackage{flushend}

\usepackage[utf8]{inputenc}
\usepackage[T1]{fontenc} 
\usepackage[russian, english]{babel}

\usepackage{epstopdf}

\usepackage[hyphens]{url} 
\usepackage[hidelinks]{hyperref} 
\usepackage{breakurl} 

\usepackage{arydshln} 

\usepackage{mathtools} 

\graphicspath{ {./pictures/} }

\newcommand{\specialcell}[2][c]{
  \begin{tabular}[#1]{@{}c@{}}#2\end{tabular}}

\newtheorem{theorem}{Theorem}[section]
\newtheorem{definition}{Definition}
\newtheorem{example}{Example}

\floatname{algorithm}{Procedure}
\renewcommand{\algorithmicrequire}{\textbf{Input:}}

\begin{document}
\title{Calculated Attributes of Synonym Sets}
\date{}

\author{
\IEEEauthorblockN{Andrew Krizhanovsky, Alexander Kirillov}
\IEEEauthorblockA{Institute of Applied Mathematical Research of the Karelian Research Centre of the Russian Academy of Sciences \\ Petrozavodsk, Karelia, Russia \\ andrew.krizhanovsky@gmail.com, kirillov@krc.karelia.ru\\}
}
\maketitle

\begin{abstract}
The goal of formalization, proposed in this paper, is to bring together, 
as near as possible, 
the theoretic linguistic problem of synonym conception 
and the computer linguistic methods based generally
on empirical intuitive unjustified factors. 
Using the word vector representation we  have  proposed the geometric 
approach to mathematical modeling of synonym set (synset).
The word embedding is based on the neural networks (Skip-gram, CBOW), 
developed and realized as word2vec  program by T. Mikolov.
The standard cosine similarity is used as the distance between word-vectors.
Several geometric characteristics of  the synset words are introduced: the  interior of synset,
the synset word rank and centrality. These notions are intended to select the most significant synset words, i.e.  the words 
which senses are the nearest to the sense of a synset.
Some experiments with proposed notions, based on RusVectores resources, are represented.
\end{abstract}

\section{Introduction}
The notion of synonym,  though it is in common use,  has no rigorous definition and is characterized by different approaches.
The descriptive definition runs as follows: the synonyms are the words expressing the same 
notion, identical  or close in the sense, differing from each other in  shades of meanings, belonging to different linguistic levels, having their own specific expressive tone.

This definition immediately raises several questions: what are the meanings of notion, sense and so on. Hence it is necessary to develop and 
introduce  a formalization, which would enable  to  use
quantitative analysis and characteristics for description of the relations between words. Such formalization is particularly significant 
in the natural language processing problems. 

In this paper,  the approach  to a synset mathematical modeling is proposed. The notion of synset (a set of synonyms) owes its occurrence to WordNet where different relations (synonymy, homonymy) are indicated between synsets but not between individual words~\cite{WordNet}. For this research the synonyms presented by Russian Wiktionary have been used.
Russian Wiktionary is a freely updated collaborative multifunctional multilingual online dictionary and thesaurus. Machine-readable Wiktionary, which we use in this paper, is regularly 
updated  
with the help of wikokit\footnote{\url{https://github.com/componavt/wikokit}} software 
on the base of Russian Wiktionary data~\cite{Krizhanovsky_Smirnov_2013}.

The authors of this paper represent the approach to  the  partial solution of the following problems:

\begin{itemize}
\item the automatic ordering of the synonyms in a synset according to the proximity of the words to the sense represented by synset;

\item the developing of mathematical tool for analysis, characterization and comparison of synsets and its experimental verification using the online-dictionary data (Russian Wiktionary);

\item  the detection, on the basis of the  developed mathematical tool (in future investigation),   of the  "weak" synsets in order to improve the dictionaries;

\item the significant problem, which has  incented  the authors to turn to this paper, 
is the word sense disambiguation~(WSD). Our main task is to combine the neural networks and the  proposed methods to solve the WSD problem at more qualitative level in comparison with existing methods~\cite{Kaushinis_2015}.
\end{itemize}

\section{The word vector representation: the brilliance and the poverty of NN-models construction by  word2vec tool}
The idea of a word representation, using neural networks (NN),   as a vector in some vector space has enjoyed wide popularity due to Skip-gram and CBOW constructions, proposed by T. Mikolov and his colleagues~\cite{Mikolov_2011, Mikolov_2012, Mikolov_2013}. The main advantage of these NN-models is their simplicity and possibility of their usage with the help of such  available instrument
as  word2vec developed also by T.Mikolov's group on the basis of text corpora. It is worth to note, from our point of view,  that the significant contribution to  this field of computer linguistics has been made by the Russian scientists -- A. Kutuzov and E. Kuzmenko, who have  developed, by the aid  of  word2vec, the NN-models for  Russian language, using several corpora.
They called the proposed tool RusVectores~\cite{Kutuzov_2015}.

The "poverty" of Mikolov's approach consists in rather confined possibilities of its applicability to finding out the meaningful pairs of semantic relations. One of the  most bright  examples of word2vec is the well-known  ($queen - woman + man \approx king$) is not supported by other expressive relations. The slightest deviations from the examples, representing satisfactory illustrations of the Mikolov's approach, lead to poor results.   
The lack of a formal justification of the Mikolov' approach
was pointed out in the recent paper of Goldberg and Levy~\cite{Goldberg_2014_word2vec}, which ends with the following appeal to researchers

\begin{quote}
        \textit{"Can we make this intuition more precise? We’d really like to see something more formal"}~\cite{Goldberg_2014_word2vec}.
\end{quote}

The presented paper, to some degree, is the partial response to this challenge of the well-known researchers in the computer linguistics.

Let us consider the main idea of the word vector representation. 
Denote by $D$ some dictionary and enumerate in some way its  words. Let  $|D|$ be the number of the words in $D$, 
$i$~--- the index number of a word in the dictionary.
\begin{definition}
The vector dictionary is the set $D=\{w_i \in \mathbb{R}^{|D|}\}$, where the $i$-th  component of  a vector  $w_i$ equals  1, while the other components are zeros.
\end{definition}
Thus, $w_i$ is the image of the $i$-th word in $D$.
The problem of the word vector representation, as it is understood at present, is to construct a linear mapping 
$L: D \rightarrow \mathbb{R}^N$,
where  $N<<|D|$, and  vector $v=L(w), w \in D$, $v$ has components  $v_j \in  \mathbb{R}$.
These procedure is called the distributed word vector representation.
Its goal is to  replace very thin set $D \in \mathbb{R}^{|D|}$, consisting of vectors with zero mutual  inner (scalar) product, by some subset of $\mathbb{R}^N$, where  $N << |D|$, with the following property: the inner products of vectors from $\mathbb{R}^N$ may be used as a measure of the words similarity, which is currently accepted in the corresponding problems of
nature language processing. If $W$ is a matrix of such linear mapping $L$ then $v=Ww$ for $v \in \mathbb{R}^N$. In addition several methods, particularly based on neural networks, are used to construct $W$. Recently, CBOW and  Skip-gram methods  has become widely used. Their mathematical basis is the modified maximum likelihood method.

For instance,  the Skip-gram NN-model provides the matrix $W$, mentioned above, as the matrix which maximizes the following function $F(W)$ $$
F(W)=\frac{1}{T}\sum_{t=1}^T \sum_{-c\leq j\leq c, j\neq 0} \ln p(w_{t+j}|w_{t})
$$
$$
p(w_{t+j}|w_t)=\frac{\exp u_{t+j}}{\sum_{i=1}^{|D|}\exp u_i}, \qquad  u_i = (Ww_i, Ww_t)
$$
where $(\cdot, \cdot)$~--- the symbol of inner product, $T$~--- the volume of training context. Here, a word $w_t$ is given in order to find out the appropriate context, containing this word and having the size $2c$ (the size of the "window").
The CBOW (continuous bag of words) NN-model, on the contrary, operates with some given context and provide an appropriate word.
These model take into account only local context. There exist some attempts to use global context (the whole document) \cite{Huang_2012}. Such approach would be useful for solving the problems of WSD.

\section{The synset geometry}

\subsection{The synset interior: IntS}

The distance between word-vectors (normalized) is measured by their inner product, i.e. by the angle between them, as in the theory of projective spaces.
Thus, the increasing of inner product corresponds to decreasing of the distance  $sim\{a, b\}$ (similarity) between vector-words $a, b \in \mathbb{R}^N$. Hence,
$sim\{a, b\} = \frac{ (a,b) }{ ||a|| \cdot ||b|| }$, where  $(a,b)$ is the inner product of vectors $a$ and $b$, $||\cdot||$ is the norm symbol. There are proposed some other 
measures of a distance between the vectors but they are based on the inner product~\cite{Levy_2015_Improving}, \cite{Mahadevan_2015_Reasoning}, \cite{Sidorov_2014_Soft}.

Let us introduce designations for normalized sum of vectors:  $M((a_{i}),n)=\frac{\sum_{i=1}^n a_{i}}{||\sum_{i=1}^n a_{i}||}$. In what follows, the distance between the sets of vectors 
will be measured by the distance between normalized mean vectors of the sets. Thus, if $A=\{a_1,...,a_n\}$ and $B=\{b_1,...,b_m\}$, $a_i, b_j \in \mathbb{R}^N$, then 
$sim\{A, B\}=(M((a_i),n),(M((b_j),m)))$.

Consider a synset $S=\{v_k, k=1,...,|S|\}$. Let us remove any word  $v$ from $S$ (the index of a word is omitted for brevity). Divide the set  $S\setminus \{v\}$ into two disjunctive subsets:  $S\setminus \{v\}=\{v_{i_s}\}\sqcup \{v_{j_p}\},$ $ s=1,...,q,$  $ p=1,...,r,$ $q+r=|S|-1, \ i_s\neq j_p $. Denote $S_1=\{v_{i_s}\}, S_2=\{v_{j_p}\}$. 
Then $S\setminus \{v\}=S_1 \cup S_2$.

\begin{definition}
The interior $Int S$ of a synset $S$ is the set of all vectors  $v \in S$ satisfying the following condition
\begin{equation}
\begin{split}
        Int S = \{v \in S:\ & sim\{S_1, S_2\} < sim\{S_1 \cup v, S_2\} \ \bigwedge \\ 
                            & sim\{S_1, S_2\} < sim\{S_1, S_2 \cup v\}\}
\end{split}
\end{equation}
for all disjunctive partitions $S\setminus \{v\}=S_1 \sqcup S_2$, 
where $S_1 \neq \varnothing,\ S_2 \neq \varnothing$.
\end{definition}
The sense of this definition: the addition of  the vector $v \in Int S$ to any of two  subsets of 
$S\setminus \{v\}$, forming its disjunctive partition, decreases the distance between these subsets (i.e. increases the similarity).  

\begin{figure}[h]
   \centering
    \includegraphics[keepaspectratio=true,width=0.98\columnwidth]{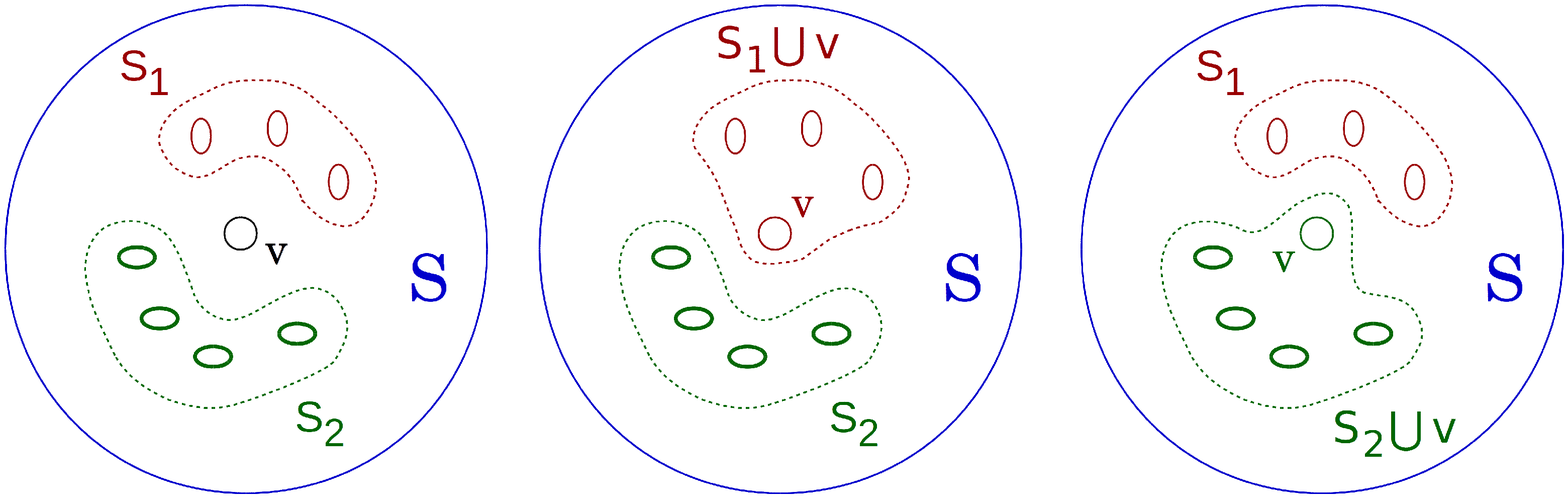}
    \caption{Vector $v$ decreases the distance between  $S_1$ and $S_2$. If it occurs for all disjunctive partitions of $S\setminus \{v\}$ then  $v \in IntS$}
    \label{fig:IntSWithTwoSets}
\end{figure}

To illustrate the notion of $IntS$, consider   two-dimensional  vectors. 
In Fig.~\ref{fig:IntSWithTwoSets} vector $v$ (conditionally shown as a circle), added to $S_1$ or $S_2$, decreases the distance between  $S_1$ and $S_2$.

\subsection{Rank and centrality of a word in synset}

\begin{figure*}[h]
   \centering
    \includegraphics[keepaspectratio=true,width=1.4\columnwidth]{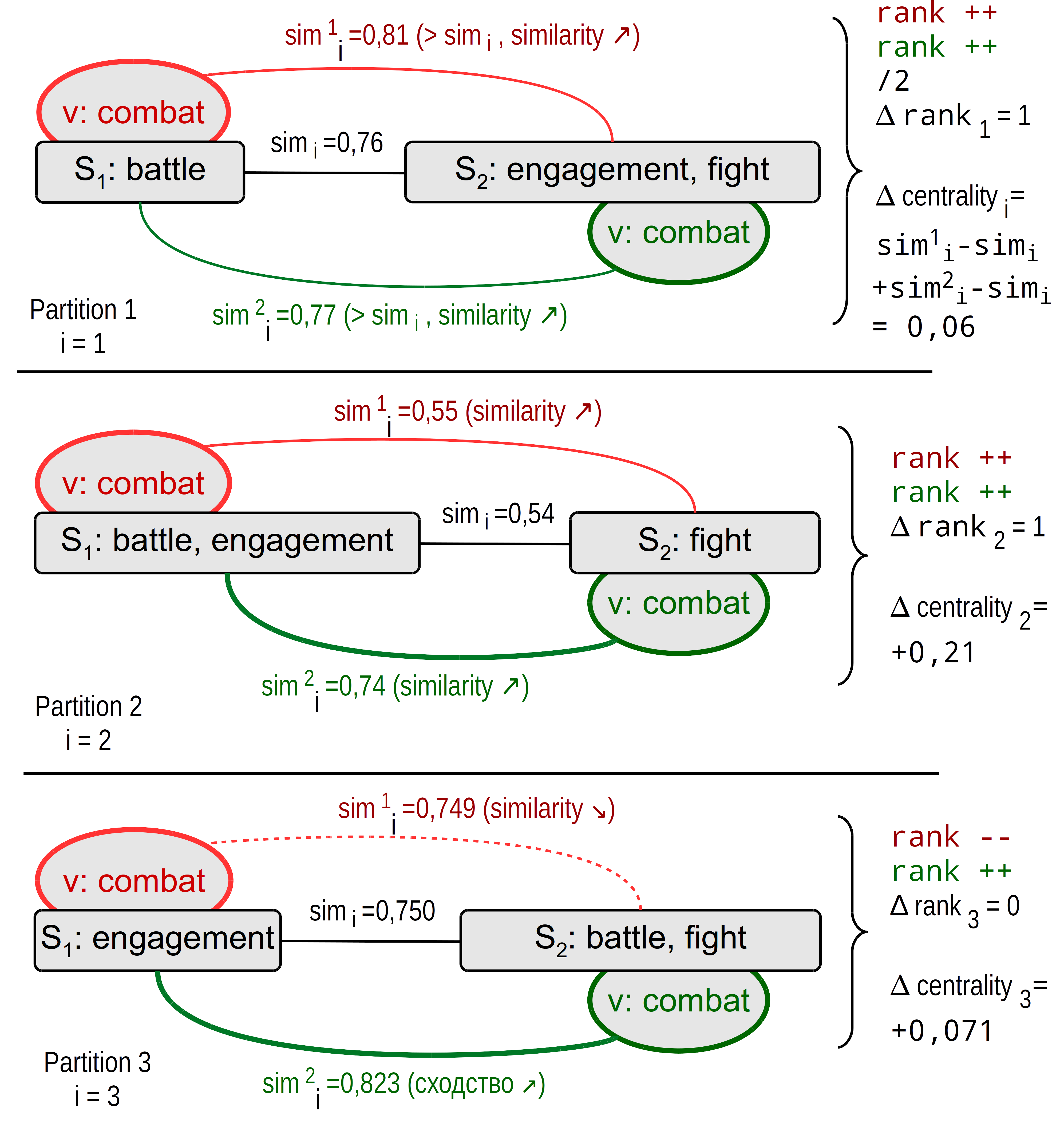}
    \caption{The values of rank and centrality for the word "combat" in synset $S=\{battle, engagement, fight, combat\}$. 
Three possible partitions of the set $S\setminus \{combat\}= \{battle, engagement, fight\}$ into two nonempty subsets, $S_1^i, S_2^i, i=1,2,3$, are presented. 
For the brevity, $S_1^i, S_2^i$ are denoted as $S_1, S_2$ respectively.
The values of $rank(v)$ and $centrality(v)$, $v=\{combat\}$ are calculated as the sums of appropriate  $\Delta rank_i := rank(v, p_i)$ and $\Delta centrality_i := centrality (v, p_i)$.}
    \label{fig:SynsetSetsRank}
\end{figure*}

Let us introduce the notion of  \textbf{the rank of a synonym} $v\in S$.
In what follows we consider only disjunctive partitions and thus, for brevity, the disjunctive partition into two subsets, the elements of partition, we shall call the partition.
Let  $P_v= \{p_i, i=1,...,2^{n-2}-1\}$~be the set of all enumerated in some way partitions  $p_i$ of the set $S\setminus\{v\}$, where $|S| = n$. Here $|S|$ is the power (the number of elements) of $S$. Suppose  $n>2$.
Consider any partition $p_i$ of the set $S \setminus \{v\}$: $S \setminus \{v\} = S_1 \sqcup S_2$.
Denote $sim_i=sim\{ S_1, S_2 \}$, 
$sim^1_i = sim\{ S_1 \cup v, S_2 \}$,
$sim^2_i = sim\{ S_1, S_2 \cup v \}$. 
Using these designations, we obtain
\begin{equation}
    \label{eqn:eq_short_ints}
    Int S = \{v \in S:  sim_i < sim^1_i \ \wedge \ sim_i < sim^2_i\}
\end{equation}
Introduce the function $r_v: P_v\rightarrow \{-1, 0, 1\}$  such that 
\begin{equation}
    \label{eqn:eq_rv_pi_definition}
r_v (p_i) = \begin{dcases} 
\ \ -1, & sim^1_i < sim_i \bigwedge sim^2_i < sim_i,\\
        & v\ \text{moving apart of}\ S_1\ \text{from}\ S_2 \\[2ex]
\ \  1, & sim^1_i > sim_i \bigwedge sim^2_i > sim_i,\\
        & v\ \text{approaching of}\ S_1\ \text{and}\ S_2 \\[2ex]
\ \  0, & (sim^1_i - sim_i) \cdot (sim^2_i - sim_i) < 0.\\
        & \text{approaching}-\text{moving apart} \end{dcases}
\end{equation}
%
The function $r_v$ is determined for each partition and gives, metaphorically speaking, the "bricks" which will below compose the rank of a synonym.
Let us briefly explain \textit{{approaching}-\text{moving apart}} line of the above definition of $r_v$.
The expression
$(sim^1_i - sim_i) \cdot (sim^2_i - sim_i) < 0$ 
is equivalent to 
$(sim^1_i < sim_i \wedge sim^2_i > sim_i) \bigvee (sim^1_i > sim_i \wedge sim^2_i < sim_i)$. 
In other words, the function $r_v (p_i)$ has the value $0$, if the adding of a word $v$ to one of the elements of a partition $p_i$ decreases (increases) the distance  $sim_i$, but the adding to another element increases (decreases), on the contrary, the distance $sim_i$. In Fig.~\ref{fig:SynsetSetsRank} this is 3 partition.

\begin{definition}
The rank of a synonym $v \in S$, where $|S| > 2,$ is the integer of the form
\begin{equation}
    \label{eqn:eq_synonym_rank}
    rank\ (v) = \sum_{i=1}^{|P_v|} r_v (p_i).
\end{equation}
\end{definition}
The definition implies that if $v \in Int S$ then $rank\ (v) = 2^{|S|-2}-1$ is the number of all nonempty disjunctive partitions of $S\setminus \{v\}$ into two subsets, where $|S\setminus\{v\}|=|S|-1$, i.e. $rank\ (v)$  has maximum and equals to the  Stirling number of the second kind: $\textstyle \lbrace{n\atop k}\rbrace = \lbrace{|S|\atop 2}\rbrace$, where $n$ is the number of elements in the set and $k$ is the number of the  subsets in a partition, here $k=2$~\cite[p.~244]{graham1994concrete}.

The relation between $IntS$  and  $rank\ (v)$ is given by the following 
\begin{theorem}[IntS theorem]
\label{IntSrem}
Assume $|S|> 2$. Then  $v \in IntS$ if and only if the rank of a word $v$  is maximal in a given synset and equals to the  Stirling number of the second kind for partition of $S$ into two nonempty  subsets, i.e.
$$
v \in IntS \Leftrightarrow
rank\ (v) = 2^{|S|-2}-1,\ \ \text{where} \ |S| \geqslant 3,
$$
\end{theorem}
\begin{IEEEproof}
\begin{multline}
    v \in IntS     
                    \mathrel{\mathop{\Leftrightarrow}^{\mathrm{(2)}}}   
    \forall p_i: Int S = \{v \in S:  sim^1_i > sim_i \\ \wedge \ sim^2_i > sim_i \} \ \ (v\ \text{approaching of}\ S_1\ \text{and}\ S_2) \ \ 
                    \mathrel{\mathop{\Leftrightarrow}^{\mathrm{(3)}}} \\ 
    \forall p_i: r_v (p_i) = 1
                    \mathrel{\mathop{\Leftrightarrow}^{\mathrm{(4)}}} \\ 
    rank\ (v) = \sum_{i=1}^{|P_v|} 1 = |P_v| = 2^{|S|-2}-1.  
\end{multline}
since $2^{|S|-2}-1$~--- is the maximal number of nonempty disjunctive partitions into two subsets. 
\end{IEEEproof}

\begin{definition}
The centrality of a synonym  $v \in S$ under a partition   $p_i$ of  $S\setminus \{v\}$ is the following value  
$$
centrality (v, p_i) = (sim^1_i(v) - sim_i) + (sim^2_i(v)-  sim_i)
$$
\end{definition}
\begin{definition}
The centrality of a synonym $v \in S$ is the following value 
$$
centrality (v)= \sum_{i=1}^{|P_v|}centrality (v, p_i)
$$
\end{definition}

\textbf{Hypothesis 1:} it is worth to note, that the word $v$, belonging to $IntS$, has the greater rank and centrality than the other words of a synset $S$. It is likely that the rank and the centrality show the measure of significance of a word in a synset, i.e. the measure of proximity of this word to the synset sense. 
Since the centrality is a  real number it  gives more precise characteristic of a word significance in a synset  than the rank which is integer (see the table~\ref{tab:CentralityRankIntS}).

\subsection{Rank and centrality computations}

The definition of centrality implies the following centrality computation Procedures \ref{alg_calcrank_one_partition} and \ref{alg_calcrank}.

\textbf{Hypothesis 2:} the more meanings has the word the less is its rank and centrality in different synsets. 

The following example and table~\ref{tab:CentralityRankIntS} support this hypothesis. 
It is worth to note that this example is not exclusive. The verification of the hypothesis on the large amount of data  is the substance of future research.

\begin{algorithm}[H]
  \caption{Computation of rank $r_v(p_i)$ and $centrality (v, p_i)$ of a word $v$ 
and a correspondent partition $p_i$ of the synset $S$}
  \label{alg_calcrank_one_partition}

  \begin{algorithmic}[1]
      \renewcommand{\algorithmicrequire}{\textbf{Input:}}
      \REQUIRE a synset $S$, a word $v \in S$  and any correspondent \\ partition $p_i$ of $S \setminus \{v\}$; 

      \renewcommand{\algorithmicrequire}{\textbf{Require:}}
      \REQUIRE $S \setminus \{v\} = S_1 \sqcup S_2$; 
  
      \ENSURE $r_v(p_i)$, $centrality (v, p_i)$.

      \STATE $sim_i \leftarrow sim\{ S_1, S_2 \}$ \vspace{1mm}
      \STATE $sim^1_i(v) \leftarrow sim\{ S_1 \cup v, S_2 \}$ // adding of a word $v$  to $S_1$ \vspace{1mm}
      \STATE $sim^2_i(v) \leftarrow sim\{ S_1, S_2 \cup v \}$  // adding of a word $v$  to $S_2$ \vspace{1mm}
      \STATE $centrality (v, p_i) \leftarrow (sim^1_i(v) - sim_i) + (sim^2_i(v) - sim_i)$ 
      \STATE           $r_v (p_i) \leftarrow 1/2 \cdot (  \text{sgn} (sim^1_i(v) - sim_i) + $ \vspace{1mm} \\ 
              $\phantom{r_v (p_i) \leftarrow 1/2 \cdot (} \text{sgn} (sim^2_i(v) - sim_i))$, \\
              $\phantom{r_v (p_i) \leftarrow 1/2 \cdot (}$ where $\text{sgn}(x) = \begin{cases} \ \ 1, & x > 0 \\ \ \ 0, & x = 0 \\ -1, & x < 0 \end{cases}$

      \RETURN $r_v(p_i)$, $centrality (v, p_i)$ 
  \end{algorithmic}
\end{algorithm}

\begin{algorithm}[H]
  \caption{Computation of $rank(v)$ and $centrality (v)$ of a word $v$ of the synset $S$}
  \label{alg_calcrank}

  \begin{algorithmic}[1]
    \renewcommand{\algorithmicrequire}{\textbf{Input:}}
    \REQUIRE a synset $S$, a word $v \in S$; 

    \ENSURE $rank (v)$, $centrality (v)$.\vspace{1mm}

    \STATE $centrality (v) \leftarrow \sum_{i=1}^{|P_v|}centrality (v, p_i)$, \vspace{1mm}
    \STATE $rank (v) \leftarrow \sum_{i=1}^{|P_v|}r_v(p_i)$.\vspace{1mm}

    \RETURN $rank (v)$, $centrality (v)$ 
  \end{algorithmic}
\end{algorithm}

\begin{example}
\normalfont

Let us consider the synset $S$ = \textit{(battle, combat, fight, engagement)}. Let us find out  $IntS$ and calculate the rank and the centrality of each word in synset.

The example of calculating  of rank and centrality of the word  "\textit{combat}" in this synset is shown in Fig.~\ref{fig:SynsetSetsRank}. 
The set of power  $|S \setminus \{v\}|= 3$ may be decomposed in three ways into two nonempty subsets. Each partition may add 1, 0 or -1  to $rank (v)$ (Fig.~\ref{fig:SynsetSetsRank}).
The values of rank and centrality equals 2 and 0,36, respectively.

In table.~\ref{tab:CentralityRankIntS} the rank, the centrality and  $IntS$ for the words of the synset are shown.

\renewcommand{\arraystretch}{1.2}
\begin{table}[h]
    \centering
    \label{tab:CentralityRankIntS}
    \caption{Rank and centrality of each word in synset, the~belongings of synonym to IntS is shown}\vspace{4pt}
    \begin{tabular}{|c|c|c|c|c|}

    \hline
    \textbf{Russian synset} & \foreignlanguage{russian}{баталия} & \foreignlanguage{russian}{бой} & \foreignlanguage{russian}{битва} & \foreignlanguage{russian}{сражение} \\
    \hline
    \textbf{Transliteration} & batálija & boj  & bítva & sražénije \\
    \hline
%
    \textbf{Translation} & \textbf{fight} & \textbf{combat} & \textbf{battle} & \textbf{engagement} \\
    \hline
    \textbf{Centrality} & -0.12    & 0.34  & 0.45  & 0.6 \\
    \hline
    \textbf{Rank}       & -3       & 2     & 3     & 3 \\
    \hline
    \textbf{IntS}       & ---      & ---   & +     & + \\
    \hline
    \multicolumn{5}{c}{\footnotesize \rule{0pt}{2em} \specialcell{Note. The precise translation of the synset' words is a rather\\difficult task. The translation serves only to illustrate the model.}} \\
    \end{tabular}
\end{table}

According to above Theorem~\ref{IntSrem}, the rank of the synonyms belonging to the synset interior, $IntS$, equals 
$$
2^{|S|-2}-1 = 2^{|4|-2}-1 = 3
$$
Table~\ref{tab:CentralityRankIntS} shows that the words "battle" and "engagement" have the largest rank (3) and centrality.
Thus,  $Int$ (battle, engagement, fight, combat) = (battle, engagement). It means that this pair, battle and engagement, is the most close in sense to all words of the synset. 
\end{example}

Rank and centrality in this example were calculated on the basis of data from Russian National Corpus.

\section{Experiments}

\renewcommand{\arraystretch}{1.2}
\begin{table*}[h]
    \centering
    \label{tab:IntSEmpty}
    \caption{Examples of synsets with empty IntS. 
    The synsets were taken from Russian Wiktionary. 
    The words in synsets are ordered by rank and centrality.
    Two Russian corpora from the project $RusVect\bar{o}r\bar{e}s$ were used
    to construct NN-models. These models were used to find out 
    $IntS$, here $OutS = S \setminus IntS$}
    \begin{tabular}{|c|c|c|c|c|}
\hline
\specialcell{Russian\\Wiktionary\\article} & Synset (from article) & $||S||$ & $||IntS||$ & Corpus \\
\hline
    \multicolumn{5}{c}{Adverb} \\ \hline

    \specialcell{beautifully\\(\foreignlanguage{russian}{прекрасно})} & 
    \specialcell{$IntS = \varnothing$,\\
                 $OutS=\{wonderfully, remarkably, excellently, perfectly, beautifully\}$} & 5 & 0 & RNC\\ \hdashline

    \specialcell{beautifully\\(\foreignlanguage{russian}{прекрасно})} &
    \specialcell{$IntS=\{perfectly, remarkably\}$,\\
                 $OutS=\{wonderfully, beautifully, excellently\}$} & 5 & 2 & News\\ \hline

    \multicolumn{5}{c}{Adjective} \\ \hline
    
    \specialcell{stony\\(\foreignlanguage{russian}{каменный})} & 
    \specialcell{$IntS = \varnothing$,\\
                 $OutS=\{stony, heartless, hard, cruel, pitiless\}$} & 5 & 0 & RNC\\ \hdashline
    
    \specialcell{stony\\(\foreignlanguage{russian}{каменный})} & 
    \specialcell{$IntS=\{pitiless\}$,\\
                 $OutS=\{stony, heartless, hard, cruel\}$} & 5 & 1 & News\\ 
\hline
    \end{tabular}
\end{table*}

In this paper  we use the NN-models, created by the authors of the project $RusVectores$~\cite{Kutuzov_2015}, namely, the model constructed on the basis of the texts of the Russian National Corpus (RNC), and the model, constructed on the basis  of the texts of the Russian news sites
(News corpus). These model are available on the site of the project $RusVectores$~\cite{Kutuzov_2015}.

The authors of $RusVectores$, A.~Kutuzov and E.~Kuzmenko, pay attention to such  peculiarities of 
RNC as hand typesetting of the texts for the corpus updating, the regulation of the different genres text relations, the small  size of the main corpus, approximately, 107 million of words (for example, the News corpus consists of 2,4 billion of tokens).

In ~\cite{Kutuzov_2015} the notion of \textit{corpus representativeness} is introduced. The sense of this notion  is the ability of the corpus to reflect (to point at) those association for a word which will be accepted by the majority of the native language speakers.
The associations generated by NN-models according to the data from RNC and Web-corpus are just used in this paper.
The problem of comparison is reduced to finding out the words which  meanings in  Web-corpus  essentially differ  from that in RNC.
Let us take into account that for each word in corpus, via NN-model, we can obtain the list of $N$  nearest words (remind: a word is a vector). Then, the result of the corpora comparison is as follows:
for more than a half of all the words (the common words in two corpora) not less than three words of the nearest ten words were the same~\cite{Kutuzov_2015}. It means that the linguistic world images, created on the basis of RNC and Internet texts, have a lot in common. But the opposite estimation is also necessary: what is the measure of discrepancy of the NN-models?     

Let us note that the notion of  \textit{corpus representativeness} acquires the new significance in view of the NN-models created on the basis of the corpus. An unbalanced sample results in excess weight of some corpus topics and as consequence to a less exact NN-model.
 
It is significant for future experiments the following observation in ~\cite{Kutuzov_2015}. For the rare
words, presented by the small set of  contexts related to this word, the associative words, generated by NN-models, will be  inexact and doubtful. 

We have conducted the experiments for testing the proposed synset model. We have used two NN-models, constructed by the authors of  $RusVectores$ on the basis of RNC and News corpus. 
While operating with NN-models we used the program gensim\footnote{\url{http://radimrehurek.com/gensim/}}, because it contains the realization of  word2vec in Python language. The gensim program is presented in~\cite{rehurek_lrec}. The authors of  $RusVectores$ used the gensim program for the NN-models constructing as well~\cite{Kutuzov_2015}.

We developed the number of scripts on the basis  of gensim for operating with NN-models, 
for calculation of the rank and the centrality and for determination of $IntS$. These script are available online\footnote{\url{https://github.com/componavt/piwidict/tree/master/lib_ext/gensim_wsd}}.

For the several thousands of synsets extracted from the Russian Wiktionary 
the rank, the centrality and $IntS$ were calculated 
on the basis of RNC and News corpus. 
The experiments have showed that the rare words in corpora have, as a rule, empty $IntS$. 
The same word in different NN-models constructed according to different corpora 
is presented by different vectors. 
The different corpora and NN-models give different word vector representations. 
Thus, the synset interior ($IntS$) for the same word could be different~(Table \ref{tab:IntSEmpty}).

 
%




\section{Conclusion}

The world of modern linguistics may be  represented, tentatively speaking, as the union of 
two domains, attracting each other but  nevertheless weakly bound nowadays: the traditional, more qualitative, and computational linguistics.  
The strict formalization of the base notions is necessary for further development of linguistics as exact science. The formal definition of such notions as \textit{word meaning},  \textit{synonymy} and others will permit to base, to the right degree, upon  the methods and algorithms of computational linguistics (corpora linguistics, neural networks,  etc.), discrete mathematics, probability theory.

In this paper we  present an approach to some formal characterizations ($IntS$, rank, centrality)  of such significant for machine-readable dictionaries and thesauri notion as the set of synonyms~--- the synset. The proposed formal tool permits to analyse  the synsets, to compare them, to determine the significance of the words in a synset. 

In future investigations we will use the developed   tool to the problem of word sense disambiguation (WSD problem).

\section*{Acknowledgment}
This paper is supported by grant 
N 18-012-00117 from the Russian Foundation for Basic Research.


\begin{thebibliography}{6}

\bibitem{Kaushinis_2015}
\textit{}T.~V.~Kaushinis, A.~N.~Kirillov, N.~I.~Kor\-zhi\-tsky, A.~A.~Kri\-zha\-novsky, A.~V.~Pilinovich, I.~A.~Sikhonina, A.~M.~Spirkova, V.~G.~Starkova, T.~V.~Stepkina, S.~S.~Tkach, Ju.~V.~Chirkova, A.~L.~Chuharev, D.~S.~Shorets, D.~Yu.~Yankevich, E.~A.~Yaryshkina, ``A review of word-sense disambiguation methods and algorithms: Introduction'', \emph{Transactions of Karelian Research Centre of Russian Academy of Science, series Mathematical Modeling and Information Technologies}, 2015, pp.~69--98, doi: 10.17076/mat135, Web: \url{http://journals.krc.karelia.ru/index.php/mathem/article/view/135}.



\bibitem{Goldberg_2014_word2vec}
\textit{}Y.~Goldberg, O.~Levy, ``word2vec explained: Deriving Mikolov et al.'s negative-sampling word-embedding method'', \emph{arXiv preprint arXiv:1402.3722}, 2014, pp.~1--5.



\bibitem{graham1994concrete}
\textit{}R.~L.~Graham, E.~K.~Donald and O.~Patashnik, \emph{Concrete mathematics}. Addison–Wesley, 1994.


\bibitem{Huang_2012}
\textit{}E.~H.~Huang, R.~Socher, C.~D.~Manning, A.~Y.~Ng, ``Improving word representations via global context and multiple word prototypes'', \emph{in Proc. of the ACL '12}, Jeju Island, Korea, 2012, pp.~873--882, Web: \url{http://dl.acm.org/citation.cfm?id=2390524.2390645}.



\bibitem{Krizhanovsky_Smirnov_2013}
\textit{}A.~A.~Kri\-zha\-novsky, A.~V.~Smirnov, ``An approach to automated construction of a general-purpose lexical ontology based on Wiktionary'', \emph{Journal of Computer and Systems Sciences International}, 2013, N 2, pp.~215--225, doi: 10.1134/S1064230713020068, Web: \url{http://scipeople.com/publication/113533/}.



\bibitem{Kutuzov_2015}
\textit{}A.~Kutuzov, E.~Kuzmenko, ``Comparing neural lexical models of a classic national corpus and a web corpus: the case for Russian'', \emph{Computational Linguistics and Intelligent Text Processing}, 2015, pp.~47--58, doi: 10.1007/978-3-319-18111-0\_4, 
Web: \url{https://www.academia.edu/11754162/Comparing_neural_lexical_models_of_a_classic_national_corpus_and_a_web_corpus_the_case_for_Russian}.



\bibitem{Levy_2015_Improving}
\textit{}O.~Levy, Y.~Goldberg, I.~Dagan, ``Improving di\-stri\-bu\-tional similarity with lessons learned from word embeddings'', \emph{Transactions of the Association for Computational Linguistics}, 2015, vol.~3, pp.~211--225.




\bibitem{Mahadevan_2015_Reasoning}
\textit{}S.~Mahadevan, S.~Chandar, ``Reasoning about linguistic regularities in word embeddings using matrix manifolds'', 
\emph{arXiv preprint arXiv:1507.07636}, 2015, pp.~1--9.


\bibitem{Mikolov_2011}
\textit{}T.~Mikolov, S.~Kombrink, L.~Burget, J.~Cernocky, S.~Khudanpur, 
``Extensions of recurrent neural network language model'', 
\emph{in Proc. of the 2011 {IEEE} International Conf. on Acoustics Speech and Signal Processing ({ICASSP})}, 2011, doi: 10.1109/icassp.2011.5947611.


\bibitem{Mikolov_2012}
\textit{}T.~Mikolov, G.~Zweig, 
``Context dependent recurrent neural network language model'',  
\emph{in Proc. of the 2012 {IEEE} Spoken Language Technology Workshop ({SLT})}, 2012, doi: 10.1109/slt.2012.6424228.


\bibitem{Mikolov_2013}
\textit{}T.~Mikolov, K.~Chen, G.~Corrado, J.~Dean, 
``Efficient estimation of word representations in vector space'', 
\emph{arXiv preprint arXiv:1301.3781}, 2013, Web: \url{http://arxiv.org/abs/1301.3781}.




\bibitem{WordNet}
\textit{}Princeton University website, What is WordNet? Web: \url{http://wordnet.princeton.edu}.




%
\bibitem{Sidorov_2014_Soft}
\textit{}G.~Sidorov, A.~Gelbukh, H.~G{\'o}mez-Adorno, D.~Pinto,  
``Soft similarity and soft cosine measure: Similarity of features in vector space model'', 
\emph{Computaci{\'o}n y Sistemas}, 2014, vol.~18, N 3, pp.~491--504, Web: \url{http://www.scielo.org.mx/pdf/cys/v18n3/v18n3a7.pdf}.





\bibitem{rehurek_lrec}
\textit{}R.~{\v R}eh{\r u}{\v r}ek, P.~Sojka, 
``Software framework for topic modelling with large corpora'', 
\emph{in Proc. of the LREC 2010 Workshop on New Challenges for NLP Frameworks}, Valletta, Malta: University of Malta, 2010, pp.~45--50, Web: \url{http://is.muni.cz/publication/884893/en}.



\end{thebibliography}
\end{document}